# Diagnostic Rule Extraction Using Neural Networks

Vitaly G. Schetinin[1] and Anatoly I. Brazhnikov[2]
Penza State University,
40, Krasnaya str., Penza, 440017, Russia
vschetinin@mail.ru [1], anibra@yahoo.com [2]

The neural networks have trained on incomplete sets that a doctor could collect. Trained neural networks have correctly classified all the presented instances. The number of intervals entered for encoding the quantitative variables is equal two. The number of features as well as the number of neurons and layers in trained neural networks was minimal. Trained neural networks are adequately represented as a set of logical formulas that more comprehensible and easy-to-understand. These formulas are as the syndrome-complexes, which may be easily tabulated and represented as a diagnostic tables that the doctors usually use. Decision rules provide the evaluations of their confidence in which interested a doctor. Conducted clinical researches have shown that diagnostic decisions produced by symbolic rules have coincided with the doctor's conclusions.

## 1 Introduction

In practice, a doctor-diagnostician applies the diagnostic rules that consist of subjective and objective features (called as symptoms) to accurately distinguish one disease ore state of the patient from others. Subjective features that reflect the complaints, the anamnesis, and the inquiry results of the patient have fuzzy, unquantitable evaluations. In contrast, objective features are the results of laboratory and tool researches that can be represented in quantitative, interval or nominal forms. A doctor interested in that confidence of diagnostic rules would be maximal. Diagnostic rules should be not only accurate but also understandable for a doctor, which wish to know how these rules work and why their usage brings the best decisions [1, 11].

For extraction and validation of diagnostic rules, a doctor must beforehand collect a representative data set involving the observations of the symptoms that occur in similar clinical cases. In practice, the data set is usually unrepresentative set because it is difficulty to collect a hundred and thousand of examples. Therefore the confidence of decision rules depends first on the size and the quality of data set a doctor classified and second on the structure of symptoms a doctor a prior suggested. In these real-world conditions, we assume that a doctor can not exactly evaluate the dividing ability or significance of each of symptoms and because we should estimate the contribution of each symptom to the decision in order to find optimal structure and parameters of desirable diagnostic rules. Further, the extracted rules are validated on the testing set. If validation results are unsatisfactory, the rules extraction process is usually repeated under changed conditions by updating classified set (e.g. removing a contradictory example a user was able to recognize accurately, and extending a feature set). The extraction process is repeated until desirable rule of required accuracy would be found.

Recently, the machine learning methods have been exploited to extract symbolic rules [2, 3, 16]. In particular, artificial neural networks have been trained to recognize the pathological states. A neural network typically consists of a number of units performing a logical function of formal neurons incorporated in a layer. The inputs to the unit in one layer have connected through weight synapses (synaptic links) with outputs of other units. Accordingly with the connectionist idea, the neural networks are fully connected and layered - an output of each unit in one layer is connected to all inputs to units in the other layer. A neural network that consists of given number of the layers, the synaptic links, and the units is trained to minimize a network error by updating its synaptic weights. A neural network trained by this way is able effectively to extract rules and decide diagnostics problems. However, the rules performed by a fully connected network are hard-to-understand because there is huge number of its synaptic links. For training neural network, it is required also to collect the representative classified data set and spend an extensive computation time [9, 10, 16, 17].

The genetic-based self-organizing methods have been used to reduce both a neural network redundancy and learning time. In particular, the Group Method of Data Handling (GMDH) of Ivakhnenko was effectively exploiting to train a polynomial multi-layered neural network of optimal complexity on incomplete training set of small size [4, 6, 7]. Below, we discuss experimental results that were obtained by self-organizing a neural network trained to extract the symbolic diagnostic rules from incomplete trained set [5, 12-15].

## 2 Problem Statement

Let us suppose that there is a training set that consists of the $n$ clinical cases that a doctor classified as two different diseases with the symptoms $x_1, ..., x_m$, where $m$ is the number of symptoms. A doctor ensures that given set of the symptoms allows him accurately to distinguish one disease from other when these pathologies are similar in all given cases. It is required to extract from this set a diagnostic rule, which can be used for accurate classifying as seen as unseen examples on training and testing set respectively.

To solve this problem, we shall use a neural network that can be represented as a compact set of the symbolic (logical) formulas, which adequately describe behavior of trained neural network [8, 18]. Such set of symbolic rules can be easily interpreted in language that is understandable to a doctor. Interpreting symbolic rules, a doctor can get comprehensible answer to the question - why he should adopt the offered decision.

In real-world tasks, the neural networks must be trained under following conditions. First, the training set may be incomplete in the cases when a doctor can not collect more than one hundred of well-classified examples. As a doctor subjectively classifies the training examples by two classes of diseases, the contradictory and incorrect examples that occur due classification errors can not be excluded. Significance of variables $x_1, ..., x_m$ that a doctor suggested is a priori unknown. A doctor can use the input variables expressed in quantitative, Boolean or nominal forms simultaneously.

Second, a doctor interested in that trained neural network has minimal architecture that is it has consisted of minimum number of the layers, the neurons and synaptic links between them. A set of symptoms as well as a network configuration should be minimal. Under these conditions, the trained neural network should have a good predictive accuracy.

Third, trained neural network should be adequately represented as a concise set of logical formulas or production rules that are widely used in medical expert systems. In simple cases, the logical formulas can be tabulated and used as diagnostic tables without a computer. In all cases, the diagnostic rules should provide an evaluation of their confidence by calculating membership of decision to each class in the range from 0 up to 1.

Note also that a user exploiting a standard-configured computer interested in that a time required for learning neural network should not exceed several seconds.

## 3 Multi-Layered Logic Neural Network

The technique we suggested satisfies to the above requests and it allowed us to synthesize a multi-layered logical neural network of optimum complexity on incomplete training set. Within framework of this technique, the well-known principles of evolutionary self-organization, external addition of Beer and adequate variety of Ashby were used for gen-erating the neural-network individuals and selecting the best from them [7]. Due to integrating these basic principles, the technique allows us to present trained neural networks as a minimal set of logic formulas $g_i(u_1, u_2)$ from two general variables $u_1, u_2$. Table 1 depicts $d = 10$ such functions, $i = 1, ..., d$.

| $u_1$ | $U_2$ | Value of function $g_i(u_1, u_2)$ | | | | | | | | |
|---|---|---|---|---|---|---|---|---|---|---|
| | | $g_0$ | $g_3$ | $g_5$ | $g_6$ | $g_7$ | $g_8$ | $g_{10}$ | $g_{12}$ | $g_{13}$ |
| 0 | 0 | 0 | 0 | 0 | 0 | 1 | 1 | 1 | 1 | 1 |
| 0 | 1 | 0 | 1 | 1 | 1 | 0 | 0 | 0 | 1 | 1 |
| 1 | 0 | 0 | 0 | 1 | 1 | 0 | 0 | 1 | 0 | 1 |
| 1 | 1 | 1 | 0 | 0 | 1 | 0 | 1 | 1 | 1 | 0 |

**Table 1:** Tabulated values of some logical functions from two arguments $u_1, u_2$

Beforehand, all quantitative and nominal input variables have to be quantized and represented as fuzzy ones. The conducted researches have shown that more effective mean to do such transformation uses some threshold $u_i$ and function $h(u_i)$ introduced to encode each variable $x_i$. The threshold $u_i$ and function $h(u_i)$ are chosen so that the number $e_i$ of classification errors that an feature $x_i$ produces on training set was minimal. Exceeding a variable $x_i$ of a threshold $u_i$ is coded by 0 or 1 in according with two kind of encoding function $h(u_i)$.

Each neural-network unit of first layer performs a logical function of formal neuron $g_i(x_j, x_k)$, $j \neq k = 1, ..., m$ consecutively for $i = 1, ..., d$. For each of these units, the classification error $e_i$ is calculated on training set. The number $e_i$ then is compared to the numbers $e_j$ and $e_k$ of classification errors that the features $x_j$ and $x_k$ produced on training set. The unit $g_i$ is removed from current layer if one of two conditions is satisfied

$$e_i > e_j \text{ or } e_i > e_k. \tag{1}$$

In contrary, value $y_j$ of this unit is used in the next layer in which a composite unit $g_i(y_j, x_k)$ is created. This procedure is repeated and each time the new layer of network is added while formal neurons satisfy to the criterion (1) of selection. In the result of self-organization, the network composed of a few layers is synthesized that consist of the best neural units.

Synthesized network can be easily presented as a set of symbolic rules *M-of-N* or syndromes $s_1..., s_N$ that form a syndrome-complex

$y = M\text{-}of\text{-}N(s_1, ..., s_N)$,
$s_i = g(x_k, x_l)$, $i = 1, ..., N$,

where *M* is number of syndromes, which is sufficient for decision making with some confidence; *N* is number of all units in last layer or number of all syndromes.

We can see that the number *M* is a level of decision making, which varies from $N_1$ to $N$, where $N_1$ is around $1/2N$. Obviously, that confidence of decision rules is increased while a number *M* growths and it will be maximal when $M = N$.

## 4 Extraction of Diagnostic Rules

The developed technique was used to extract the differential diagnostics rules from incomplete classified sets of history cases that experienced doctors can carefully collect. In all cases, the concise sets of logical formulas were found which were presented as the diagnostic tables. The extracted diagnostic rules are represented below.

**4.1 Diagnosing Infectious Endocarditis (IE) and System Red lupus (SRL)**

Neural network was trained on the set consisted of 18 cases of IE and SRL, encoded as $y = 0$ and $y = 1$ respectively. A priori, the inputs to neural network were 24 syndromes, seven of which were laboratory (quantitative). The extracted diagnostic rule includes only two laboratory variables and six clinical features that listed in Table 2.

| Features | | $u_i$ | $h(u_i)$ |
|---|---|---|---|
| 1. Leucocytes, $10^9$/L | $x_2$ | 6.2 | 0 |
| 2. circulatting immune complex, opt. units | $x_5$ | 130.0 | 0 |
| 3. Articular syndrome | $x_8$ | | |
| 4. Anhelation | $x_{11}$ | | |
| 5. Erythema of skin | $x_{13}$ | | |
| 6. noises in heart | $x_{14}$ | | |
| 7. Hepatomegaly | $x_{15}$ | | |
| 8. Myocarditis | $x_{16}$ | | |

**Table 2:** The features for diagnosing infectious endocarditis and system red lupus

In Table 2, the thresholds $u_i$ and encoding functions $h(u_i)$ also shown. The rest 16 features that doctors offered were rejected as uninformative ones.

In symbolic form, the trained neural network can be adequately represented as syndrome-complex consisting of $N = 9$ syndromes

$$y = M\text{-of-}N(\ g_6(y_{125}, x_2),\ g_0(y_{39}, x_8),\ g_0(y_{41}, x_8),\ g_0(y_{42}, x_8),\ g_0(y_{111}, x_8),$$
$$g_0(y_{124}, x_8),\ g_0(y_{131}, x_8),\ g_{10}(y_{73}, x_{14}),\ g_{10}(y_{125}, x_{14})\ ),$$

where $y_i$ is variable in first layer:

$y_{39} = g_{12}(x_{11}, x_2)$, $y_{41} = g_6(x_{13}, x_2)$, $y_{42} = g_{12}(x_{14}, x_2)$, $y_{73} = g_0(x_{16}, x_5)$,
$y_{111} = g_{13}(x_{14}, x_{11})$, $y_{124} = g_{12}(x_{14}, x_{13})$, $y_{125} = g_3(x_{15}, x_{13})$,
$y_{131} = g_{10}(x_{16}, 14)$.

Note that in this case, a level of a decision making is $M = 5..., 9$. This rule is easily tabulated and it can be represented as the diagnostic Table 3.

| Features | | | $z_8$ | 0 | 1 | 0 | 1 | 0 | 1 | 0 | 1 | 0 | 1 | 0 | 1 | 0 | 1 | 0 | 1 |
|---|---|---|---|---|---|---|---|---|---|---|---|---|---|---|---|---|---|---|---|
| | | | $z_7$ | 0 | 0 | 1 | 1 | 0 | 0 | 1 | 1 | 0 | 0 | 1 | 1 | 0 | 0 | 1 | 1 |
| | | | $z_6$ | 0 | 0 | 0 | 0 | 1 | 1 | 1 | 1 | 0 | 0 | 0 | 0 | 1 | 1 | 1 | 1 |
| | | | $z_5$ | 0 | 0 | 0 | 0 | 0 | 0 | 0 | 0 | 1 | 1 | 1 | 1 | 1 | 1 | 1 | 1 |
| $z_1$ | $z_2$ | $z_3$ | $z_4$ | | | | | | | | | | | | | | | | |
| 0 | 0 | 0 | 0 | +7 | +7 | +7 | +7 | +9 | +9 | +9 | +9 | +6 | +6 | +7 | +7 | +7 | +7 | +9 | +9 |
| 0 | 0 | 0 | 1 | +7 | +7 | +7 | +7 | +9 | +9 | +9 | +9 | +6 | +6 | +7 | +7 | +7 | +7 | +9 | +9 |
| 0 | 0 | 1 | 0 | -7 | -7 | -7 | -7 | +7 | +6 | +7 | +6 | -9 | -9 | -8 | -8 | -6 | -7 | +5 | -5 |
| 0 | 0 | 1 | 1 | -6 | -6 | -6 | -6 | +9 | +8 | +9 | +8 | -8 | -8 | -7 | -7 | +5 | -5 | +7 | +6 |
| 0 | 1 | 0 | 0 | +7 | +7 | +7 | +7 | +9 | +8 | +9 | +8 | +6 | +6 | +7 | +7 | +7 | +6 | +9 | +8 |
| 0 | 1 | 0 | 1 | +7 | +7 | +7 | +7 | +9 | +8 | +9 | +8 | +6 | +6 | +7 | +7 | +7 | +6 | +9 | +8 |
| 0 | 1 | 1 | 0 | -7 | -7 | -7 | -7 | +7 | +5 | +7 | +5 | -9 | -9 | -8 | -8 | -6 | -8 | +5 | -6 |
| 0 | 1 | 1 | 1 | -6 | -6 | -6 | -6 | +9 | +7 | +9 | +7 | -8 | -8 | -7 | -7 | +5 | -6 | +7 | +5 |
| 1 | 0 | 0 | 0 | +6 | +6 | +6 | +6 | +8 | +8 | +8 | +8 | +6 | +6 | +6 | +6 | +7 | +7 | +8 | +8 |
| 1 | 0 | 0 | 1 | +6 | +6 | +6 | +6 | +8 | +8 | +8 | +8 | +6 | +6 | +6 | +6 | +7 | +7 | +8 | +8 |
| 1 | 0 | 1 | 0 | -9 | -9 | -9 | -9 | -5 | -6 | -5 | -6 | -9 | -9 | -9 | -9 | -7 | -8 | -6 | -7 |
| 1 | 0 | 1 | 1 | -9 | -9 | -9 | -9 | +5 | -5 | +5 | -5 | -9 | -9 | -9 | -9 | -6 | -7 | -5 | -6 |
| 1 | 1 | 0 | 0 | +6 | +6 | +6 | +6 | +8 | +7 | +8 | +7 | +6 | +6 | +6 | +6 | +7 | +6 | +8 | +7 |
| 1 | 1 | 0 | 1 | +6 | +6 | +6 | +6 | +8 | +7 | +8 | +7 | +6 | +6 | +6 | +6 | +7 | +6 | +8 | +7 |
| 1 | 1 | 1 | 0 | -9 | -9 | -9 | -9 | -5 | -7 | -5 | -7 | -9 | -9 | -9 | -9 | -7 | -9 | -6 | -8 |
| 1 | 1 | 1 | 1 | -9 | -9 | -9 | -9 | +5 | -6 | +5 | -6 | -9 | -9 | -9 | -9 | -6 | -8 | -5 | -7 |

**Table 3:** Diagnostic table for differing infectious endocarditis and system red lupus. If sign is plus, $+M$, then infectious endocarditis. If sign is minus, $-M$, then system red lupus. $N = 9$ is number of syndromes.

The diagnostic decisions locate in intersection of one of $2^4$ lines, that were formed by combining fuzzy features $z_1$, $z_2$, $z_3$, $z_4$, and a column, that was defined by features $z_5$, $z_6$, $z_7$, $z_8$. For example,

    **IF** *leukocytes* are less than 6.2 ($z_1$= 1), **AND**
       *circulating immune complex* is less than 130 ($z_2$= 1), **AND**
       *articular syndrome* is absent ($z_3$= 0), **AND**
       *anhelation* is absent ($z_4$= 0), **AND**
       *erythema* is absent ($z_5$= 0), **AND**
       no *noises in heart* ($z_6$= 0), **AND**
       hepatomegaly ($z_7$= 0) is absent, **AND**
       *myocarditis* is absent ($z_8$= 0),
    **THEN**
       the diagnosis is *infectious endocarditis*, (+6), with 6 syndromes from 9.

This rule correctly classifies all training examples and it was used for diagnostics more 120 clinical cases. Among unseen examples, no divergences between produced decisions and doctor conclusions exist [5, 13, 15].

**4.2. Diagnosing Infectious Endocarditis (IE) and Active Rheumatism (AR)**

Neural network was trained on set that consisted of 18 cases of IE and 17 AR, encoded by 0 and 1 respectively. A priori, the inputs to neural network were 24 syndromes as above. The diagnostic rule was found which includes one laboratory variable (*rheumatoid factor* $x_9$) and five clinical features (*articular syndrome* $x_{10}$, *headaches* $x_{12}$, *hurried pulse* $x_{19}$, *nephritis* $x_{20}$ and *pleurisy* $x_{22}$).

Trained neural network contains $r = 4$ layers that can be represented by a set of the logical formulas indicated in Table 4.

| $r = 1$ | | | | $r = 2$ | | | | $r = 3$ | | | | $r = 4$ | | | |
|---|---|---|---|---|---|---|---|---|---|---|---|---|---|---|---|
| $y^1_i = g_j(x_k, x_l)$ | | | | $y^2_i = g_j(y^1_k, x_l)$ | | | | $y^3_i = g_j(y^2_k, x_l)$ | | | | $y^4_i = g_j(y^3_k, x_l)$ | | | |
| i | j | k | l | i | j | k | l | i | j | k | l | i | j | k | l |
| 27 | 6 | 10 | 9 | 31 | 6 | 30 | 10 | 30 | 6 | 42 | 10 | 1 | 0 | 30 | 12 |
| 30 | 6 | 20 | 9 | 35 | 0 | 27 | 12 | 31 | 6 | 43 | 10 | 2 | 0 | 31 | 12 |
| | | | | 42 | 1 | 30 | 19 | 34 | 6 | 53 | 10 | 3 | 0 | 34 | 12 |
| | | | | 43 | 5 | 30 | 19 | 39 | 0 | 31 | 12 | 4 | 0 | 54 | 12 |
| | | | | 45 | 6 | 27 | 20 | 46 | 0 | 45 | 12 | 5 | 0 | 55 | 12 |
| | | | | 53 | 1 | 30 | 22 | 54 | 1 | 31 | 19 | 6 | 0 | 60 | 12 |
| | | | | | | | | 55 | 5 | 31 | 19 | 7 | 0 | 61 | 12 |
| | | | | | | | | 60 | 1 | 45 | 19 | 8 | 0 | 81 | 12 |
| | | | | | | | | 61 | 5 | 45 | 19 | 9 | 0 | 85 | 12 |
| | | | | | | | | 72 | 6 | 35 | 20 | 10 | 1 | 39 | 19 |
| | | | | | | | | 81 | 1 | 31 | 22 | 11 | 5 | 39 | 19 |
| | | | | | | | | 85 | 1 | 45 | 22 | 12 | 1 | 46 | 19 |
| | | | | | | | | | | | | 13 | 5 | 46 | 19 |
| | | | | | | | | | | | | 14 | 1 | 72 | 19 |
| | | | | | | | | | | | | 15 | 5 | 72 | 19 |
| | | | | | | | | | | | | 16 | 1 | 39 | 22 |
| | | | | | | | | | | | | 17 | 1 | 46 | 22 |
| | | | | | | | | | | | | 18 | 1 | 72 | 22 |

    **Table 4:** A set of logical formulas for diagnosing infectious endocarditis and active rheumatism

In the latter, $r = 4$, layer, $N = 18$ syndromes are represented, and consequently, the level of decision making ranges from 10 to 18. Table 5 depicts the values of diagnostic rule calculated accordingly with found set of formulas.

| Features |       | $z_6$ | 0   | 1   | 0   | 1   | 0   | 1   | 0   | 1   |
|          |       | $z_5$ | 0   | 0   | 1   | 1   | 0   | 0   | 1   | 1   |
|          |       | $z_4$ | 0   | 0   | 0   | 0   | 1   | 1   | 1   | 1   |
| $z_1$    | $z_2$ | $z_3$ |     |     |     |     |     |     |     |     |
| 0 | 0 | 0 | +18 | +18 | +15 | +16 | +15 | +15 | +15 | +16 |
| 0 | 0 | 1 | +18 | +18 | -18 | -12 | +12 | +12 | +12 | +18 |
| 0 | 1 | 0 | +18 | +18 | +15 | +16 | +15 | +15 | +15 | +16 |
| 0 | 1 | 1 | -18 | -13 | -18 | -13 | +10 | +15 | +10 | +15 |
| 1 | 0 | 0 | +18 | +18 | +15 | +16 | +15 | +15 | +15 | +16 |
| 1 | 0 | 1 | -18 | -12 | -18 | -12 | +12 | +18 | +12 | +18 |
| 1 | 1 | 0 | +18 | +18 | +15 | +16 | +15 | +15 | +15 | +16 |
| 1 | 1 | 1 | -18 | -13 | -18 | -13 | +10 | +15 | +10 | +15 |

**Table 5:** Diagnostic table for differing infectious endocarditis and active rheumatism. ($+M$) indicates infectious endocarditis, ($-M$) indicates active rheumatism. $N= 18$ is number of syndromes

Note that in Table 5 there are no contradictory situations when, for example, the number of syndromes with value 0 is equal to number of syndromes with value 1. Extracted rule correctly classified all training examples and was applied for diagnosing more than 60 clinical cases [5, 15].

### 4.3 Predicting Post-Operational Complications

To predict the post-operational complications in abdominal surgery, the neural network was trained on classified set consisted of five examples of complicated events and eight normal events, which were encoded by 0 and 1 respectively. A priori, the inputs to neural network were 19 laboratory and clinical syndromes that are typically explored before operation. Most significant features are variables $x_0, ..., x_{10}$ represented in Table 6.

| Features | | $u_i$ | $h(u_i)$ |
|---|---|---|---|
| 1. Expected duration of operation | $x_0$ | 4.3 | 0 |
| 2. Hemoglobin, gram/l | $x_1$ | 90.9 | 1 |
| 3. Erythrocytes, $10^{12}$ | $x_2$ | 3.3 | 1 |
| 4. Speed of erythrocyte subside, millimeter/hour | $x_3$ | 18.0 | 0 |
| 5. Residual nitrogen, micromole/l | $x_4$ | 21.4 | 0 |
| 6. Sugar, millimole/l | $x_5$ | 4.6 | 0 |
| 7. Bilirubin total, micromole/l | $x_6$ | 15.4 | 0 |
| 8. Urea (Carbamide), millimole/l | $x_7$ | 6.5 | 0 |
| 9. Albumen, gram/l | $x_8$ | 63.7 | 1 |
| 10. Fibrinogen, gram/l | $x_9$ | 1.3 | 1 |
| 11. Protein index, % | $x_{10}$ | 70.6 | 1 |

**Table 6:** Initial features for predicting complications

Trained neural network accurately classified all training examples. It consists of two layers, last layer is consisted of $N = 22$ neurons. The inputs to neural network are only seven variables that are $x_3$, $x_4$, $x_5$, $x_6$, $x_8$, $x_9$, $x_{10}$. Trained neural network can be represented as a set of logical formulas represented in Table 7.

| $r = 1$ | | | | $r = 2$ | | | |
|---|---|---|---|---|---|---|---|
| $y^1_i = g_j(x_k, x_l)$ | | | | $y^2_i = g_j(y^1_k, x_l)$ | | | |
| $i$ | $j$ | $k$ | $l$ | $i$ | $j$ | $k$ | $l$ |
| 147 | 0 | 10 | 4 | 1 | 0 | 191 | 6 |
| 148 | 8 | 10 | 4 | 2 | 8 | 191 | 6 |
| 149 | 0 | 10 | 6 | 3 | 0 | 188 | 9 |
| 152 | 0 | 10 | 8 | 4 | 8 | 188 | 9 |
| 153 | 8 | 10 | 8 | 5 | 0 | 188 | 10 |
| 161 | 8 | 9 | 4 | 6 | 0 | 177 | 10 |
| 162 | 0 | 9 | 6 | 7 | 8 | 177 | 10 |
| 163 | 8 | 9 | 6 | 8 | 0 | 163 | 4 |
| 177 | 8 | 8 | 4 | 9 | 0 | 162 | 4 |
| 188 | 0 | 6 | 4 | 10 | 0 | 161 | 6 |
| 191 | 6 | 5 | 3 | 11 | 0 | 160 | 6 |
| | | | | 12 | 0 | 153 | 4 |
| | | | | 13 | 8 | 153 | 4 |
| | | | | 14 | 0 | 152 | 4 |
| | | | | 15 | 8 | 152 | 4 |
| | | | | 16 | 0 | 149 | 4 |
| | | | | 17 | 0 | 148 | 6 |
| | | | | 18 | 0 | 148 | 8 |
| | | | | 19 | 8 | 148 | 8 |
| | | | | 20 | 0 | 147 | 6 |
| | | | | 21 | 0 | 147 | 8 |
| | | | | 22 | 8 | 147 | 8 |

**Table 7:** A set of logical formulas for predicting complications

Extracted diagnostic rule was tested on set consisted of 118 clinical events, which were evaluated by a few doctors. On this set, the error rate was 12%. As additional researches shown, most errors occur because of the doctors were using different criteria to evaluate and classify the complications of their patients [14].

## 5 Conclusions

In all considered cases, the neural network were trained on incomplete sets of classified instances that experienced doctors could suggest. Trained neural networks correctly classified the presented examples. At that, the number of intervals entered for encoding quantitative variables was minimal, equal to two. The number of features as well as the number of neurons and layers in the trained neural networks is minimal. Trained neural networks were adequately represented in symbolic form that is easy to understanding. One simple form is similar to a syndrome-complex that the doctors use typically. Other is the set of logical formulas that can be easily tabulated and interpreted to be useful for diagnostic goals. The decision rules provide the evaluations of their confidence in which interested a doctor. The conducted clinical verification has shown that most decisions that symbolic rules produced have coincided with the doctor's conclusions.